\documentclass[10pt, a4paper, logo, twocolumn, copyright]{psibot}

\usepackage{times}
\usepackage[numbers]{natbib}
\usepackage{multicol}

\usepackage{times}
\usepackage{amsmath}
\usepackage{amssymb}

\usepackage{tcolorbox}
\usepackage{colortbl}
\usepackage{geometry}

\usepackage[numbers]{natbib}
\usepackage{multicol}
\usepackage{graphicx}
\usepackage{booktabs}
\usepackage{multirow}
\usepackage{subcaption}
\usepackage{float} 
\usepackage{cuted}
\usepackage{cleveref}
\usepackage{siunitx}
\usepackage{caption}
\usepackage{boldline}
\usepackage{listings}
\usepackage{parskip}

\usepackage{hyperref}

\usepackage{microtype}
\usepackage{graphicx}
\usepackage{amsmath}  
\usepackage{amssymb}  
\usepackage{booktabs} 
\usepackage{hyperref}
\usepackage{multirow} 
\usepackage{colortbl}  
\usepackage{booktabs}  
\usepackage{subcaption}
\usepackage{lipsum}
\usepackage{xurl}
\definecolor{lightblue}{RGB}{173, 216, 230}
\definecolor{red}{RGB}{205,0,0}
\newcommand{\tbfred}[1]{\textbf{\textcolor{red}{#1}}}

\newcommand{\wall}{\tbfred{Wall}}
\newcommand{\edge}{\tbfred{Edge}}
\usepackage{hyperref}

\definecolor{pmcolor}{RGB}{70,70,70}
\newcommand{\pmcolor}[1]{\textcolor{pmcolor}{#1}}

\newcommand{\pms}[2]{#1{\tiny{{\pmcolor{{$\pm$#2}}}}}}

\usepackage{amsmath}
\usepackage{amssymb}
\usepackage{mathtools}
\usepackage{amsthm}

\theoremstyle{plain}

\theoremstyle{definition}

\theoremstyle{remark}

\usepackage[textsize=tiny]{todonotes}

\begin{document}
 
\title{Dexterous Non-Prehensile Manipulation for Ungraspable Object via Extrinsic Dexterity}

\author[1,2]{Yuhan Wang}
\author[1,2]{Yu Li}
\author[1,2,\dag]{Yaodong Yang}
\author[1,2,\dag]{Yuanpei Chen}
\correspondingauthor{Yuanpei Chen\{yuanpei.chen312@gmail.com\},Yaodong Yang\{yaodong.yang@pku.edu.cn\}}

\affil[1]{Peking University}
\affil[2]{PKU-PsiBot Joint Lab}



%

\maketitle

\begin{strip}
    \centering
    \includegraphics[width=1.0\linewidth]{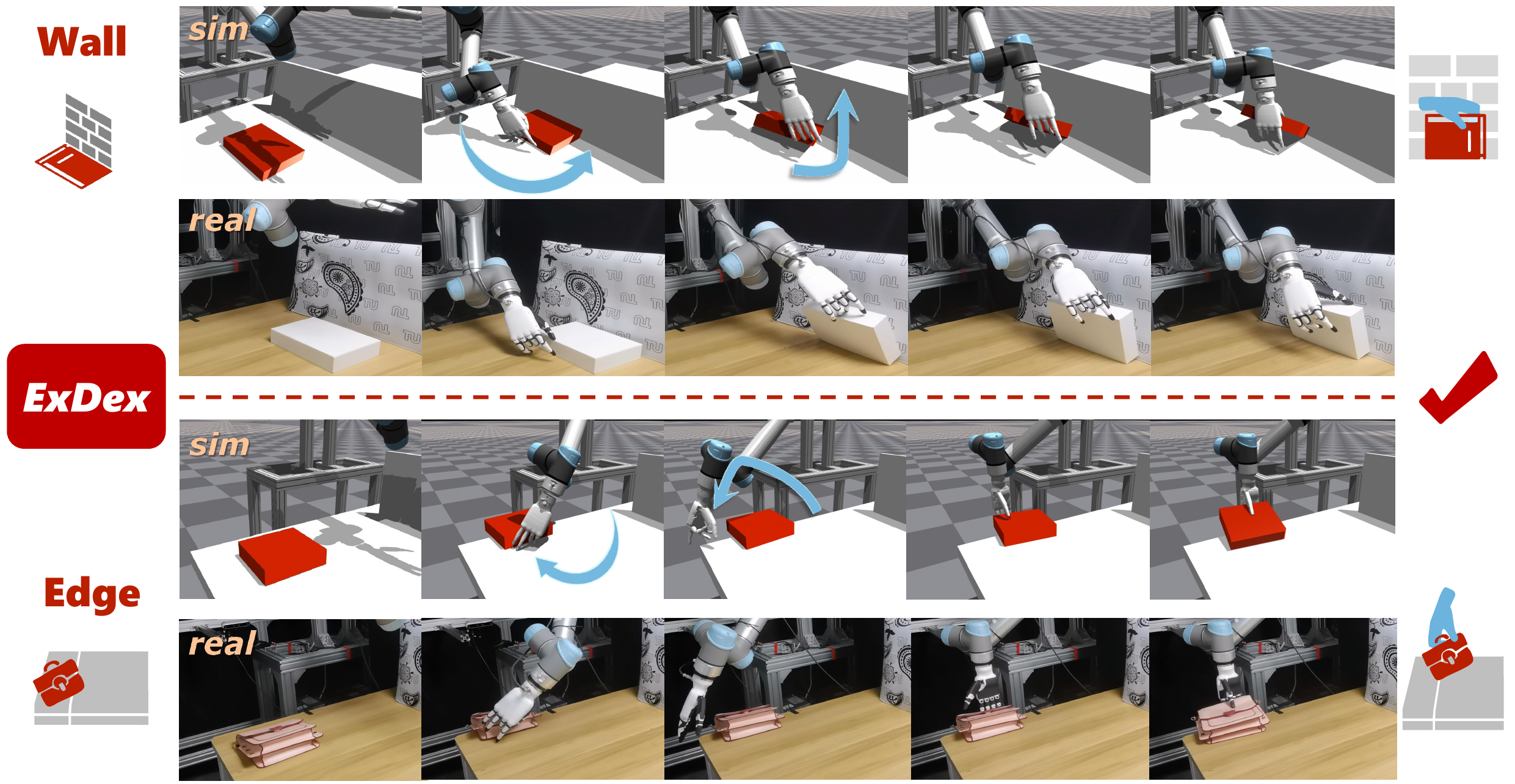}
    \captionof{figure}{We present \textbf{ExDex}, a hierarchical framework that enables non-prehensile manipulation skills for ungraspable objects leveraging external environments. Our approach is demonstrated through two representative tasks: \wall, where objects are pushed against walls to facilitate manipulation. And \edge, where objects are repositioned to table edges allowing the hand to maneuver into optimal grasping poses. Through reinforcement learning in simulation, we successfully train these manipulation policies and achieve zero-shot transfer to real-world scenarios.}
    \label{fig:teaser}
\end{strip}

\begin{abstract}
Objects with large base areas become ungraspable when they exceed the end-effector's maximum aperture. Existing approaches address this limitation through extrinsic dexterity, which exploits environmental features for non-prehensile manipulation. While grippers have shown some success in this domain, dexterous hands offer superior flexibility and manipulation capabilities that enable richer environmental interactions, though they present greater control challenges. Here we present ExDex, a dexterous arm-hand system that leverages reinforcement learning to enable non-prehensile manipulation for grasping ungraspable objects. Our system learns two strategic manipulation sequences: relocating objects from table centers to edges for direct grasping, or to walls where extrinsic dexterity enables grasping through environmental interaction. We validate our approach through extensive experiments with dozens of diverse household objects, demonstrating both superior performance and generalization capabilities with novel objects. Furthermore, we successfully transfer the learned policies from simulation to a real-world robot system without additional training, further demonstrating its applicability in real-world scenarios. Project website: \url{https://tangty11.github.io/ExDex/}.

\end{abstract}

\vspace{-15pt}

\section{Introduction}

Humans naturally manipulate objects with their multi-finger dexterous hands through a rich repertoire of strategies. Beyond direct grasping, humans demonstrate remarkable ability to exploit environmental features for manipulation. For instance, when encountering large, flat objects placed in the middle of a table that are challenging to grasp directly, humans intuitively leverage environmental constraints like walls or table edges. They seamlessly combine non-prehensile actions such as pushing, sliding, and pivoting with dexterous manipulation to achieve reliable grasps. This adaptive exploitation of environmental affordances enables humans to handle objects that would otherwise be ungraspable through direct manipulation alone. Such environment-aware manipulation strategies significantly expand the range of objects that can be successfully manipulated, demonstrating the sophisticated interplay between dexterous control and environmental interaction.

Replicating human-like extrinsic dexterity in multi-finger robotic hands remains an unexplored yet crucial problem in robotics. Traditional approaches to dexterous manipulation primarily rely on trajectory optimization with simplified contact models~\cite{chen2024springgrasp, mordatch2012contact}. However, these methods often fail in contact-rich scenarios due to the complexity of modeling dynamic contact interactions and the uncertainty in physical parameters. While imitation learning has demonstrated promising results in direct dexterous manipulation tasks~\cite{chen2022learning, shaw2023videodex, wang2024dexcap}, it faces significant limitations when applied to extrinsic dexterity. The collection of high-quality demonstration data through human teleoperation becomes particularly challenging for dynamic contact-rich manipulations, as operators struggle to precisely control multiple fingers while maintaining stable environmental contacts. 
Recent years have witnessed remarkable progress in applying reinforcement learning to robotic systems~\cite{openai2019solving, yang2024anyrotate, pitz2023dextrous, handa2022dextreme}. Leveraging large-scale parallel simulation, reinforcement learning enables robots to undergo extensive training in simulation and then deploy the learned policies to the real-world. This approach offers several key advantages for learning extrinsic dexterity skills: First, parallel simulation environments allow for rapid policy exploration without concerns about physical robot wear or safety constraints. Second, we can systematically vary object properties and environmental conditions during training, leading to more robust policies. Third, the framework naturally handles contact-rich scenarios without requiring explicit contact modeling or expert demonstrations. These benefits make reinforcement learning particularly suitable for learning complex manipulation strategies that combine dexterous control with environmental interactions.

Unlike previous works that train end-to-end reinforcement learning policies~\cite{chen2021system}, relocating objects from table centers to edges or walls requires a more sophisticated approach. This task demands a sequence of non-prehensile manipulation skills and strategic planning. While existing research often simplifies the problem by placing objects near external contacts~\cite{zhou2023learning, chen2023synthesizing}, they overlook the complexity of strategic object repositioning and environmental interaction.
Even training individual manipulation skills through reinforcement learning presents significant challenges. The high-dimensional action space of multi-finger dexterous hands, combined with the contact-rich nature of environmental interactions, makes each subtask difficult to learn. Beyond these difficulties, the challenge lies in high-level strategic planning. Optimal object relocation requires consideration of multiple factors: the current object position, robot arm configuration, and kinematic constraints. Simply choosing the nearest environmental contact point is insufficient and often leads to suboptimal or failed manipulations. Instead, the system must evaluate potential target locations while considering the robot's reachability, joint limits, and possible collision-free paths. This comprehensive planning approach significantly improves manipulation success rates compared to naive nearest-neighbor strategies.

Here we present \textbf{ExDex} a framework for dexterous manipulation of ungraspable objects using extrinsic dexterity with multi-finger hands, focusing particularly on leveraging table edges and walls. We introduce a hierarchical learning approach combining a high-level planner for identifying optimal environmental contacts with a low-level controller for precise non-prehensile manipulation. The high-level planner generates target poses and transition signals, while the low-level controller executes pushing policies to achieve these poses, followed by object retrieval under external contacts.
The experiments in both simulation and real-world settings validate our framework's effectiveness. Our results demonstrate successful generalization to unseen objects and zero-shot transfer to physical systems.

In summary, our main contribution encompasses:
\begin{itemize}
    \item First exploration of extrinsic dexterity with multi-finger dexterous manipulation in both simulation and real-world scenarios.
    \item Novel hierarchical framework combining high-level planning and low-level control for occluded grasp tasks.
    \item Extensive experimental validation demonstrating system effectiveness across simulated and physical environments.
\end{itemize}

\section{Related Works}
\subsection{Dexterous Manipulation}

Multi-finger dexterous manipulation remains a significant challenge in robotics. Traditional approaches attempt to solve this through trajectory optimization based on dynamic models~\cite{chen2024springgrasp, mordatch2012contact, bai2014dexterous, kumar2014real}. However, these models often rely on simplified contact assumptions, failing when confronted with complex, contact-rich tasks. While~\cite{chen2023synthesizing} proposed combining graph search, optimal control, and learning-based objective functions for extrinsic dexterity in pre-grasp operations, their results were limited to simulation, with some predicted trajectories proving impractical for physical robot execution.
Recent research has demonstrated remarkable success with imitation learning~\cite{mandikal2021learning, chen2022learning, shaw2023videodex, wang2024dexcap, chen2022dextransfer, rajeswaran2017learning, radosavovic2021state, arunachalam2023holo, guzey2023dexterity, handa2020dexpilot, sivakumar2022robotic, qin2023anyteleop, qin2022dexmv, cui2022play, haldar2023teach, qin2022one, arunachalam2022dexterous, guzey2023see, lin2024learning}. 3D-ViTac~\cite{huang20243dvitac} achieves precise manipulation using tactile feedback, while DexCap~\cite{wang2024dexcap} enables complex bimanual tasks through in-the-wild data collection using data gloves. However, imitation learning's reliance on human demonstration data presents substantial costs. Moreover, teleoperation latency makes collecting data for highly dynamic actions particularly challenging (such as flipping objects from wall edges).

In recent years, reinforcement learning (RL) has been widely adopted for dexterous hand manipulation tasks, spanning in-hand object reorientation~\cite{chen2021system, chen2022visual, yin2023rotating, qi2023general, qi2023hand, dasari2023learning, openai2019solving, yang2024anyrotate, pitz2023dextrous, handa2022dextreme, khandate2023sampling}, bimanual manipulation~\cite{huang2023dynamic, lin2024twisting, chen2022towards}, pre-grasping~\cite{zhou2023learning, ding2024preafford}, hand long-horizon manipulation~\cite{chen2023sequential, huang2023dynamic}. Dynamic Handover~\cite{huang2023dynamic} demonstrated multi-agent reinforcement learning for dynamic ball-catching between dexterous hands, while Robopianist~\cite{zakka2023robopianist} achieved human-level piano playing through simulation-based training.
We develop policies that adapt to dynamic motion control of real robots using RL. To our knowledge, this work represents the first exploration of extrinsic dexterity with dexterous hands demonstrated in both simulation and real-world environments.

\subsection{Extrinsic Dexterity}

External environmental resources such as contacts, gravity, and dynamic motions\cite{dafle2014extrinsic} enable robot hands to grasp and manipulate objects even without suitable contact points. Previous work has demonstrated the utility of environmental interactions, including external contacts for object grasping\cite{zhou2023learning, ma2024dexdiff, ding2024preafford, chen2023synthesizing} and reorientation~\cite{stepputtis2018extrinsic}, as well as leveraging gravity~\cite{dong2023robotic} or dynamic motions~\cite{ha2022flingbot, dafle2014extrinsic} to improve grasping postures. However, except for \cite{chen2023synthesizing}, these works primarily employ grippers or underactuated multi-finger hands. We instead utilize a five-fingered dexterous hand, leveraging its greater degrees of freedom and flexibility for enhanced grasping capabilities and improved policy generalization across multiple objects.

Zhou et al.~\cite{zhou2023learning} present the closest approach to ours, learning a closed-loop policy through reinforcement learning. However, their work relies on restrictive assumptions, including objects being initially positioned near walls and walls being sufficiently low for grippers to access objects from above. In contrast, our approach accommodates objects anywhere in the workspace. PreAfford~\cite{ding2024preafford} learns affordance to guide object-centric interaction point selection for extrinsic dexterity and object repositioning. However, their reliance on motion planning or assumed availability of low-level controllers becomes problematic in complex environments or with challenging objects. Moreover, most previous work has not explored how to leverage the high degrees of freedom of dexterous hands for extrinsic dexterity. While UniDexFPM~\cite{wu2024unidexfpm} investigated dexterous hand pre-grasp manipulation in tabletop environments, their results were limited to simulation without physical robot validation.

We first relocate objects to locations where external conditions can be effectively utilized, such as beside walls or table edges, through non-prehensile manipulation (pushing) before leveraging these conditions for object retrieval. To address the challenge of insertion of fingers between walls and objects after pushing, we learn a strategy that maintains appropriate spacing when positioning objects against walls.

\vspace{-5pt}

\begin{figure*}[t]
    \centering
    \includegraphics[width=0.9\linewidth]{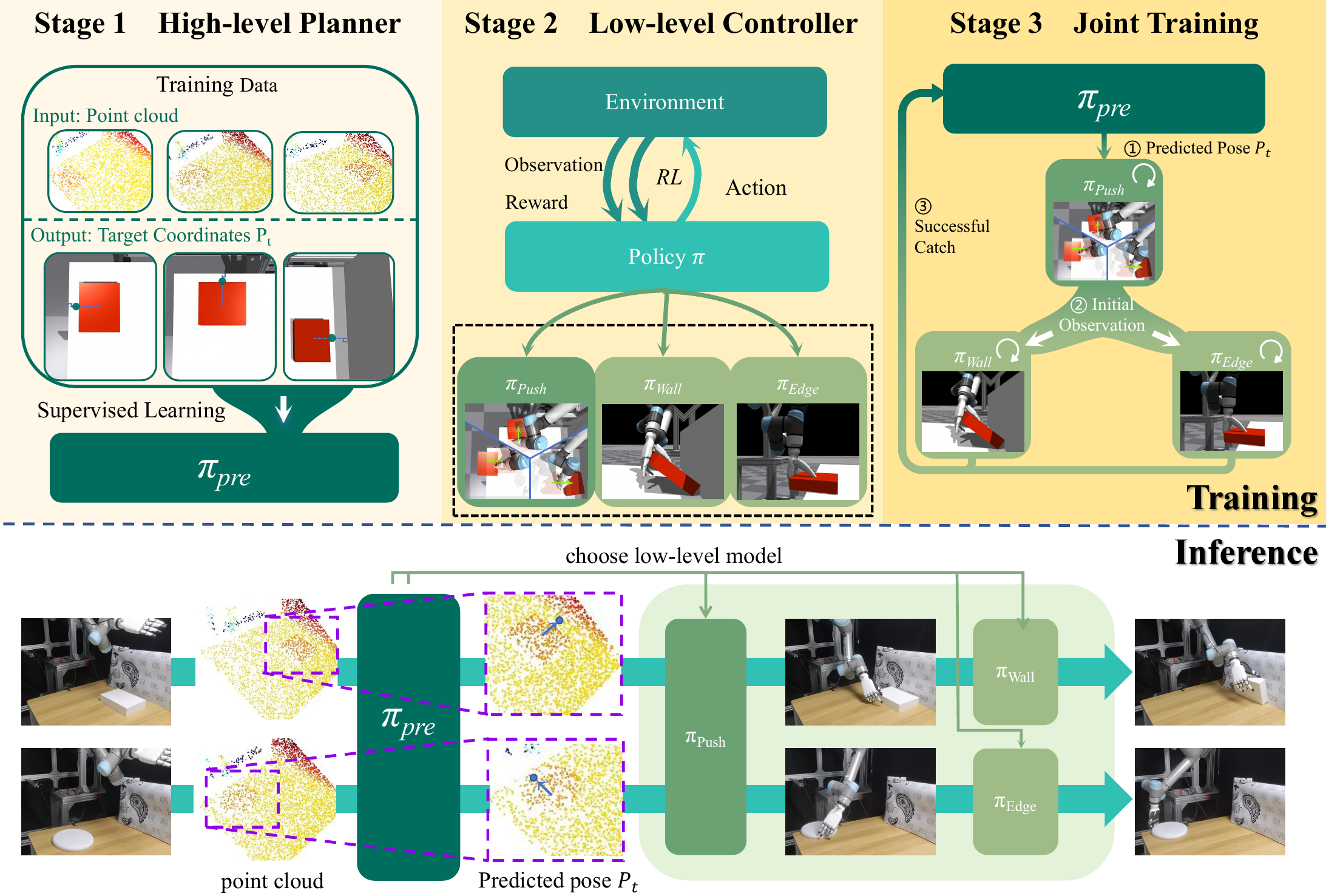}
    \caption{\textbf{Illustration of the ExDex System Design.} (A) Training: Our system is trained in three stages. In Stage 1, we train a prediction model $\pi_\textrm{pre}$ through supervised learning that takes point cloud input and predicts the optimal target position $P_t$ for object repositioning. Stage 2 focuses on training three low-level skills via reinforcement learning: a pushing policy $\pi_\textrm{push}$ that repositions objects to target locations, and two policies $\pi_\textrm{wall}, \pi_\textrm{edge}$ that enable grasping of ungraspable objects from walls and table edges via extrinsic dexterity. In Stage 3, we jointly finetune these policies to ensure better transitions between consecutive skills. (B) Inference: During inference, our system first use the $\pi_\textrm{pre}$ to process the environmental point cloud to determine whether to execute the $\pi_\textrm{wall}$ or $\pi_\textrm{edge}$, while simultaneously predicting the corresponding target position $P_t$. The pushing policy $\pi_\textrm{push}$ then moves the object to this target position, followed by the selected extrinsic dexterity policy ($\pi_{\textrm{wall}}$ or $\pi_{\textrm{edge}}$) to complete the grasp.}  
    \label{fig:method}
\end{figure*}

\section{Task Formulation} 
\label{sec:formulation}

In this paper, we address the challenge of grasping ungraspable objects that having large, flat base surfaces using a dexterous multi-finger hand. The task objective is to employ non-prehensile manipulation to reposition objects near environmental features that can assist in successful grasping. We formulate this task as a finite horizon Markov Decision Process (MDP), defined by the 5-tuple ($\mathcal{S}, \mathcal{A}, R, P, \gamma$). Here, $\mathcal{S}$ and $\mathcal{A}$ represent the state and action spaces respectively. The stochastic dynamics $P:\mathcal{S} \times \mathcal{A} \times \mathcal{S} \rightarrow [0, 1]$ determine the probability of transitioning to state $s^\prime$ given current state $s$ and action $a$. $R: \mathcal{S} \times \mathcal{A} \times \mathcal{S} \rightarrow \mathbb{R}$ defines the reward function, and $\gamma \in (0,1)$ is the discount factor. Our objective is to train a policy $\pi$ that maximizes the expected cumulative reward $\mathbb{E}_{\pi}[\sum_{t=0}^{T-1}\gamma^{t}R]$ in an episode with $T$ time steps.

\vspace{-15pt}

\section{Method} 
\label{sec:method}

In this section, we introduce our system for dexterous non-prehensile manipulation for ungraspable object. The overview of the system is shown in Figure~\ref{fig:method}. Our framework consists of three parts: the High-level Planner Design (Section~\ref{sec:high_level}), Low-level Policy Training (Section~\ref{sec:rl_design}) and Joint Training(Section~\ref{sec:joint_training}). The details of our sim-to-real policy transfer are introduced in Section~\ref{sec:sim2real}.

\subsection{High-level Planner Design}\label{sec:high_level}

The first step in extrinsic dexterity is relocating objects to environments that can be leveraged for manipulation, such as walls or table edges. Therefore, planning a desired location where external conditions can be effectively utilized is crucial for successful extrinsic dexterity. To achieve this, we train a prediction model $\pi_{pre}$ to predict target positions for object relocation. The model takes environmental point cloud data $p$ as input and outputs three-dimensional target coordinates $P_t=(P_x, P_y, P_z)_t$ and a signal $s$ to pick a low-level policy automatically for subsequent applying. The predicted $P_t$ serves as the target position for the $\pi_{push}$ policy in the low-level controller.



\subsection{Low-level Policy Training}\label{sec:rl_design}

Based on the predicted target pose $P_t$ from the high-level planner, we train three specialized policies using model-free reinforcement learning: (1) A $\pi_{\textrm{push}}$ policy that pushes objects to the target position $P_t$; (2) A $\pi_{\textrm{wall}}$ policy for grasping objects near walls starting from $P_t$; and (3) A $\pi_{\textrm{edge}}$ policy for retrieving objects from table edges at $P_t$. The following subsections detail the observation and action space, reward design, and training methodology.

\noindent \textbf{Observation Space.}
The observation space 
\begin{equation}
\mathcal{S} = \{q_t, \{F_t^{f, i}\}_{i=1}^{5}, p_t^\textrm{obj}, v_t^\textrm{obj}, P_t, c_p\}
\end{equation}
consists of several components: robot state (proprioceptive arm and hand joint positions $q_t \in \mathbb{R}^{18}$, five fingertip poses $\{F_t^{f, i}\}_{i=1}^{5}\in \mathbb{R}^{15}$), object pose $p_t^\textrm{obj}\in SE(3)$ and velocity $v_t^\textrm{obj}\in \mathbb{R}^{6}$, target information (predicted pose $P_t$ from high-level planner), and contact information (hand-designed contact position $c_p \in \mathbb{R}^{3}$ that maintains a fixed relative position to the object center).

\noindent \textbf{Action Space.}
The action space $a_t = \{a_t^{\text{arm}}, a_t^{\text{hand}}\}$ consists of two components: hand joint positions $a_t^{\text{hand}} \in \mathbb{R}^{6}$ and relative arm joint positions $a_t^{\text{arm}} \in \mathbb{R}^{6}$. For the hand, the policy directly outputs absolute joint angles $a_t^{\text{hand}}$ as target positions. For the arm, the policy generates relative position changes $a_t^{\text{arm}}$, which are added to the current joint angles to obtain target positions. The PD controller then converts these target positions into joint torques for both the arm and hand.

\noindent \textbf{Reward Design.}
To reduce the complexity of reward shaping, we unify our reward function into three components with a staged reward mechanism. The next stage reward is only calculated when specific conditions are met. Specifically:

\begin{equation}
r = r_{\textrm{motion}} + r_{\textrm{pregrasp}} \cdot P(a) + r_{\textrm{grasp}} \cdot P(b)
\label{equation:rew}
\end{equation}

where $P(\cdot)$ represents condition probabilities. 
In the following, we describe each reward component in detail. All reward terms share the same goal of minimizing distances between their arguments, thus we denote these proximity-based functions as $T(\cdot,\cdot)$, which output larger values as their arguments become closer. The specific implementation of $P(\cdot)$, $T(\cdot,\cdot)$ and hyperparameter can be found in Appendix~\ref{app:reward_design}.

\subsubsection{Motion reward $r_{\textrm{motion}}$}
The motion reward is designed to guide either object movement to a target pose or fingertip positioning for manipulation.
For $\pi_{\textrm{push}}$ and $\pi_{\textrm{wall}}$, it encourages the object to reach specific target poses: $r_{\textrm{motion}}=T(P_t^{obj}, P_t^{target})$. In $\pi_{\textrm{push}}$, $P_t^{target}$ is set to the pose $P_t$ predicted by the high-level planner, while in $\pi_{\textrm{wall}}$, $P_t^{target}$ is a pre-defined pose above the object to facilitate extrinsic dexterity.
For $\pi_{\textrm{edge}}$, the reward guides fingertip positioning: $r_{\textrm{motion}} = T(\{F_t^{f, i}\}_{i=1}^{5}, P_t^{target})$, encouraging the thumb to position above the object while placing the other four fingers beneath the object.

\subsubsection{Pre-grasp reward $r_{\textrm{pregrasp}}$}
The pre-grasp reward encourages the hand to achieve an advantageous pre-grasp pose after object repositioning: $r_{\textrm{pregrasp}}=T(F_t^{f,3}, c_p)$, where $F_t^{f,3}$ is the position of the middle fingertip, and $c_p$ is a relative fixed point on the object.

\subsubsection{Grasp reward $r_{\textrm{grasp}}$}
Once in pre-grasp position, the grasp reward promotes stable grasping by optimizing the collective positions of all five fingertips relative to the object: $r_{\textrm{grasp}}=T(P_t^m, p_t^{obj})$, where $P_t^m = \frac{F_t^{f,1} + F_t^{f,2}}{3}$ represents the midpoint between thumb ($F_t^{f,1}$) and middle fingertip ($F_t^{f,3}$) positions.

\noindent \textbf{Training Method}
We employ PPO~\cite{schulman2017proximal} to train our low-level control policies, leveraging its stability and sample efficiency. The training process is accelerated through parallel simulations in IsaacGym, enabling simultaneous training across 4096 environments. To improve policy robustness, we incorporate comprehensive domain randomization techniques, including variations in robot and object properties.
Furthermore, we adopt a curriculum learning strategy to enhance training efficiency. The training begins with some similar objects (Figure~\ref{fig:setup} (a-pretrain)) at a fixed initial position. As the success rate improves, we gradually increase task complexity by introducing objects with larger difference in size(Figure~\ref{fig:setup} (a-finetune)) and randomizing their initial positions. This progressive learning approach helps the policies develop robust manipulation skills while maintaining stable training dynamics.

\subsection{Joint Finetuning}\label{sec:joint_training}

Sequentially executing trained policies often leads to poor performance in long-horizon manipulation tasks. This is primarily because the terminal state of previous policy may not align well with the initial state distribution of the subsequent policy, creating a state distribution mismatch that affects policy execution. This challenge, known as the skill chaining problem~\cite{chen2023sequential, konidaris2009skill, sutton1999between}, requires special consideration in policy training.
To address this issue, we jointly finetune our policies with the following order. 
Firstly, to enhance the robustness of $\pi_{\text{push}}$ against potential biases in the high-level planner's predictions, we introduce Gaussian noise with a specified standard deviation to the predicted target pose. 
Then, to better align the transition states between sequential non-prehensile skills, we replace the initial states of both $\pi_{\text{wall}}$ and $\pi_{\text{edge}}$ with terminal states obtained from $\pi_{\text{push}}$ rollouts. 
Finally, to improve the capability of the high level planner $\pi_{\text{pre}}$ to predict the target pose with a higher likelihood of success, we refine it with the target poses from successful rollouts for each task.
This approach ensures smoother transitions between different manipulation phases, enhancing the overall task performance.


\subsection{Sim-to-Real Transfer}\label{sec:sim2real}
When applying the RL policy to the realworld, some environment states can not be estimated accurately like joint and object velocity. Therefore, we follow the teacher-student distilling framework~\cite{chen2022visual, chen2021system} to zero-shot transfer our simulation policy into the real dexterous arm-hand system. 
Specifically, we rollout our policies in simulation sequentially to collect the whole teacher demonstration trajectories. For distillation from demonstration, we employ a transformer-based imitation learning network to predict the target arm-hand joint angles. To mitigate the observation gap between simulation and real-world, the distilled student policy only takes the low dimension state observation including proprioception and object 6d pose as input. 
To obtain object 6d pose, we use Segment Anything model\cite{kirillov2023segment} to get the initial mask of the object, followed by FoundationPose\cite{wen2023foundationpose} for pose estimation and tracking. 
We also build a digital twin framework for sim-to-real transfer. More details about the teacher-student distillation and digital twin can be found in Appendix~\ref{app:sim2real}.


\begin{figure}[!t]
    \centering
    \includegraphics[width=0.95\linewidth]{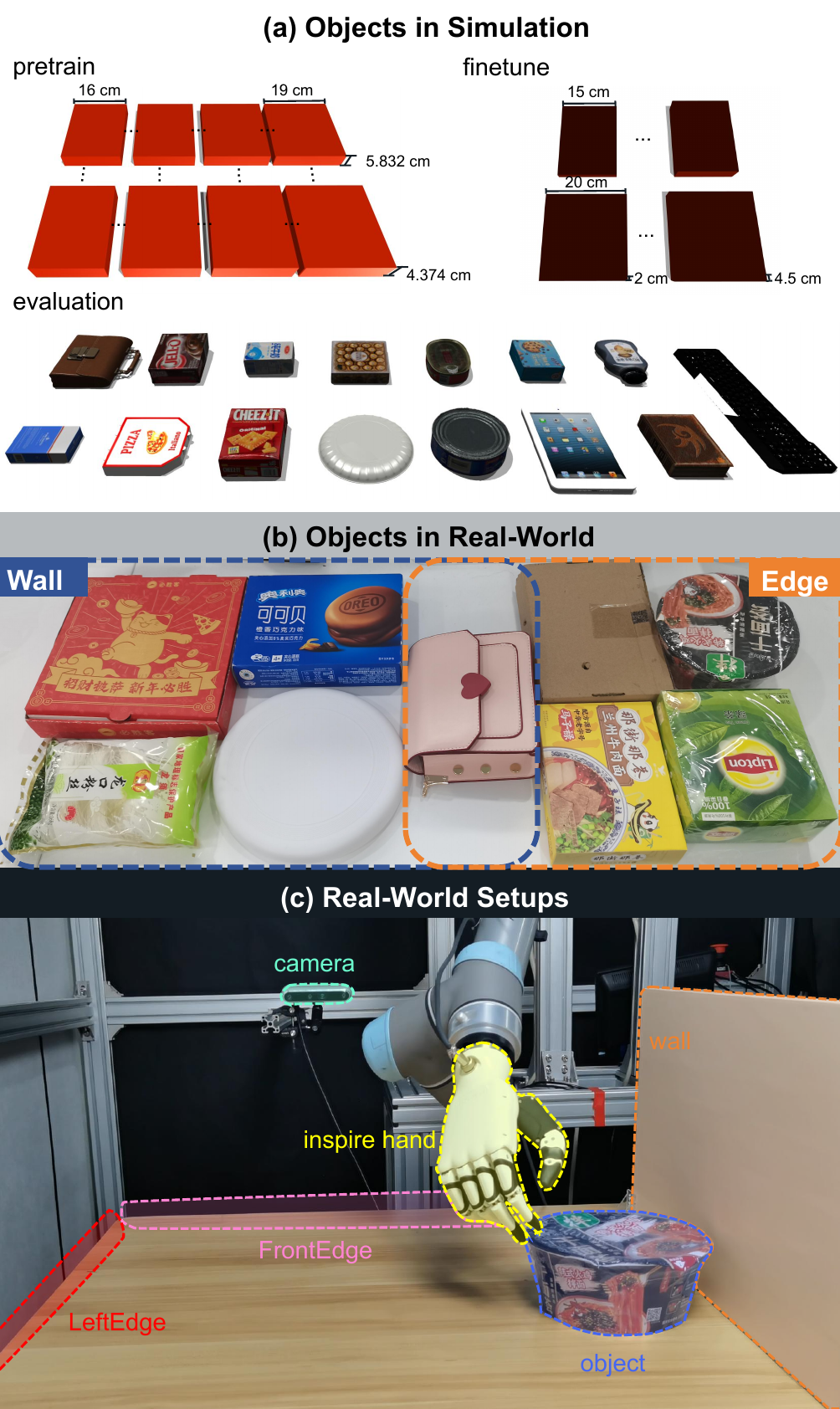}
    \caption{\textbf{Overview of the environment setups.} (a) Object sets used in simulation. Policies are firstly trained on the pretrain set, and then finetuned on the finetune set, and tested for zero-shot generalization on the unseen set. (b) Real-world test objects (top: wall-task objects, bottom: edge-task objects). (c) Workspace of the real-world, We use an Inspired Hand mounted on a UR5e robot, equipped with a RealSense D455 camera.}  
    \label{fig:setup}
\end{figure}

\section{Experiment} 
\label{sec:exp}

In this section, we comprehensively evaluate the performance of our proposed framework in simulation and real world to address the following questions:

(1) Can our high-level planner generate optimal object relocating strategy given different external environments? (Section~\ref{subsec:high-level})

(2) Is the dexterous hand motion learned by our low-level polices necessary for our tasks? (Section~\ref{subsec:dexterous})
    
(3) Is our reward design suitable for the non-prehensile manipulation skill training? (Section~\ref{subsec:reward})
    
(4) Can our joint finetuning strategy improve the generalizability and robustness of our framework? (Section~\ref{subsec:joint})
    
(5) Can our framework learned in simulation be applied to a real-world dexterous arm-hand robot system? (Section~\ref{subsec:real-world})

First, we introduce the main setup of our experiments including our dataset, evaluation metric and several baselines for comparison with our method. Then we evaluate the effectiveness of our framework separately from high-level and low-level parts through quantitative and qualitative results. Finally, we provide details of how we conduct real-world experiments and the performance of our method.
Our simulation and real-world settings are shown in Figure~\ref{fig:setup}. All simulation results in tables are evaluated in 3 different seeds, and the real-world results are evaluated in 10 trials for each object.

\subsection{Setup}
\textbf{Dataset} 
In simulation environments, we only use box with various physics property ratios as the training asset. We evaluate the generalizability of our framework on other 15 objects with diverse geometries. In real-world scenario, we select 10 objects with different sizes and physics properties for sim-to-real evaluation. The objects used in simulation and real-world are shown in Figure~\ref{fig:setup}.

\textbf{Evaluation and Metric}\label{exp:SR}
We evaluate the performance of our framework in a scene as shown in Figure~\ref{fig:setup}(c) containing three external contacts\textemdash Wall, FrontEdge and LeftEdge\textemdash each representating a separate task. For all tasks, objects are initially placed randomly in the center of the table. We train individual policies for each task as follow. For the Wall task, the policy $\pi_{wall}$ is trained to first push the object to a wall-adjacent position before utilizing the wall to assist in grasping. For the FrontEdge and LeftEdge task, we follow the training paradigm of $\pi_{edge}$, where the object is pushed to the respective table edge prior to maneuvering the hand to an appropriate position for grasping.
We introduce the following metrics for evaluating the performance:
(1) \textit{Target Transition Error} (TTE) is the Euclidean distance in centimeters between the target position predicted by the high-level planner and ground truth.
(2) \textit{Success Rate} (SR) is the percentage of successful grasping after a series of non-prehensile manipulation. We define success if the object is grasped steadily above a height threshold (10 cm).
These metrics evaluate different aspects of our framework: TTE assesses the accuracy of the high-level planner, and SR evaluates the overall task completion success.

\textbf{Baseline}
We compare our methods with the following baselines and ablations.
(1) \textit{Random Target.} This baseline maintains our low-level policies but replaces the high-level planner with random target position selection. The random positions are constrained to theoretically feasible regions (e.g., ensuring partial object overhang for the FrontEdge and LeftEdge tasks, or positions near walls for the Wall task) to maintain basic task feasibility.
(2) \textit{Arm-Only.} Our low-level policies trained with arm control only. The hand joint angle is fixed.
(3) \textit{Heuristic.} Using predefined action primitives as low-level policies.
(4) \textit{Ours w/o MR.} Our low-level policies trained without motion reward.
(5) \textit{Ours w/o ST.} Our low-level policies trained without stage reward mechanism.
(6) \textit{Ours w/o JH.} Our low-level policies without joint finetuning for the high-level planner.
(7) \textit{Ours w/o JL.} Our low-level policies without joint finetuning for the low-level policies.

\subsection{High-level planner}
\label{subsec:high-level}
\begin{table*}[!t]
\caption{
   Quantitative comparison of the high-level planner. 
   }
\label{tab:high_level}
\centering
\scriptsize
\setlength{\tabcolsep}{1pt}
\begin{tabular}{l|cc|cc|cc|cc|cc|cc}

\toprule
 & \multicolumn{4}{c|}{Wall} & \multicolumn{4}{c|}{Edge} & \multicolumn{4}{c}{Left Edge}\\

\midrule
\multirow{2}{*}{Method}
 & \multicolumn{2}{c|}{Seen}
 & \multicolumn{2}{c|}{Unseen}
 & \multicolumn{2}{c|}{Seen}
 & \multicolumn{2}{c|}{Unseen}
 & \multicolumn{2}{c|}{Seen}
 & \multicolumn{2}{c}{Unseen}\\

 & TTE  & SR
 & TTE  & SR
 & TTE  & SR
 & TTE  & SR 
 & TTE  & SR
 & TTE  & SR 
 \\
\midrule
Random Target 
& \pms{45.19}{1.94}  & \pms{66.21}{0.96}
& \pms{45.15}{1.90}  & \pms{54.50}{1.47}
& \pms{31.63}{2.77}  & \pms{88.23}{1.04}
& \pms{31.75}{2.87}  & \pms{60.11}{2.21}
& \pms{31.23}{1.23}  & \pms{74.28}{0.12}
& \pms{31.32}{1.24}  & \textbf{\pms{57.22}{2.70}}
\\

\textbf{Ours} 
& \textbf{ \pms{0.41}{0.00}}  & \textbf{ \pms{83.25}{0.34}} 
& \textbf{ \pms{0.41}{0.01}}  & \textbf{ \pms{54.94}{1.57}} 
& \textbf{ \pms{3.16}{0.03}}  & \textbf{ \pms{89.43}{0.68}}
& \textbf{ \pms{2.85}{0.06}}  & \textbf{ \pms{68.00}{3.55}}
& \textbf{ \pms{2.58}{0.01}}  & \textbf{ \pms{76.75}{2.41}} 
& \textbf{ \pms{2.97}{0.07}}  & \pms{54.39}{2.27}

\\
\bottomrule

\end{tabular}

\end{table*}
To evaluate the generalizability of the high-level planner for different external contacts, we compare the relocating strategy of our high-level planner with \textit{Random Target}. The quantitative results in Table~\ref{tab:high_level} shows that our high-level planner generates superior relocating positions with lower target transition error (TTE) across all tasks and objects, which facilitates the continual non-prehensile manipulation skills with higher success rate (SR). These results demonstrate that our high-level planner can accurately predict target positions suitable for all the tasks. (add more analysis for different task)Moreover, the improved SR indicates that after joint finetuning, our prediction model tends to select target positions where subsequent policies have higher success rates, suggesting effective integration between the high-level planner and low-level controller.

\subsection{Low-level controller}
\begin{table*}[!ht]
\caption{
   Quantitative comparison of the low-level controller. 
   }
\label{tab:baseline_SR}
\centering
\setlength{\tabcolsep}{7pt}

\begin{tabular}{l|cc|cc|cc}

\toprule

\multirow{2}{*}{Method}
 & \multicolumn{2}{c|}{Wall}
 & \multicolumn{2}{c|}{FrontEdge}
 & \multicolumn{2}{c}{LeftEdge}\\

 & Seen & Unseen  & Seen & Unseen  & Seen & Unseen \\

\midrule

  



ArmOnly. 
& \pms{16.27}{0.18}  & \pms{2.50}{1.22}
& \pms{29.95}{0.25}  & \pms{26.39}{2.74} 
& \pms{14.28}{2.39}  & \pms{19.61}{0.70} 
\\

Heuristic. 
& \pms{8.03}{0.30}  & \pms{2.72}{0.61} 
& \pms{73.70}{0.61}  & \pms{56.61}{1.86} 
& \pms{63.91}{1.22}  & \pms{48.61}{0.98} 
\\

w/o MR. 
& \pms{0.00}{0.00}  & \pms{0.17}{0.14}
& \pms{0.00}{0.00}  & \pms{1.00}{0.14}
& \pms{0.00}{0.00}  & \pms{0.01}{0.02}
\\

w/o ST. 
& \pms{0.01}{0.02}  & \pms{0.22}{0.21}
& \pms{2.28}{0.24}  & \pms{5.44}{0.42}
& \pms{0.56}{0.31}  & \pms{1.36}{0.14} 
\\

w/o JL. 
& \pms{49.47}{3.67}  & \pms{19.39}{2.91}
& \pms{74.56}{3.31}  & \pms{32.61}{2.21} 
& \pms{48.18}{0.47}  & \pms{12.94}{0.55} 
\\

w/o JH. 
& \textbf{\pms{83.88}{0.33}}  & \pms{54.28}{0.97} 
& \pms{82.02}{0.92}  & \pms{63.22}{3.80} 
& \pms{73.33}{2.70}  & \pms{53.44}{3.91} 
\\

\textbf{Ours} 
& \pms{83.25}{0.34}  & \textbf{\pms{54.94}{1.57}}
& \textbf{\pms{89.43}{0.68}}  & \textbf{\pms{68.00}{3.55}}
& \textbf{\pms{76.75}{2.41}}  & \textbf{\pms{54.39}{2.27}}
\\
\bottomrule

\end{tabular}

\end{table*}

\begin{figure}[!t]
    \centering
    \includegraphics[width=1.0\linewidth]{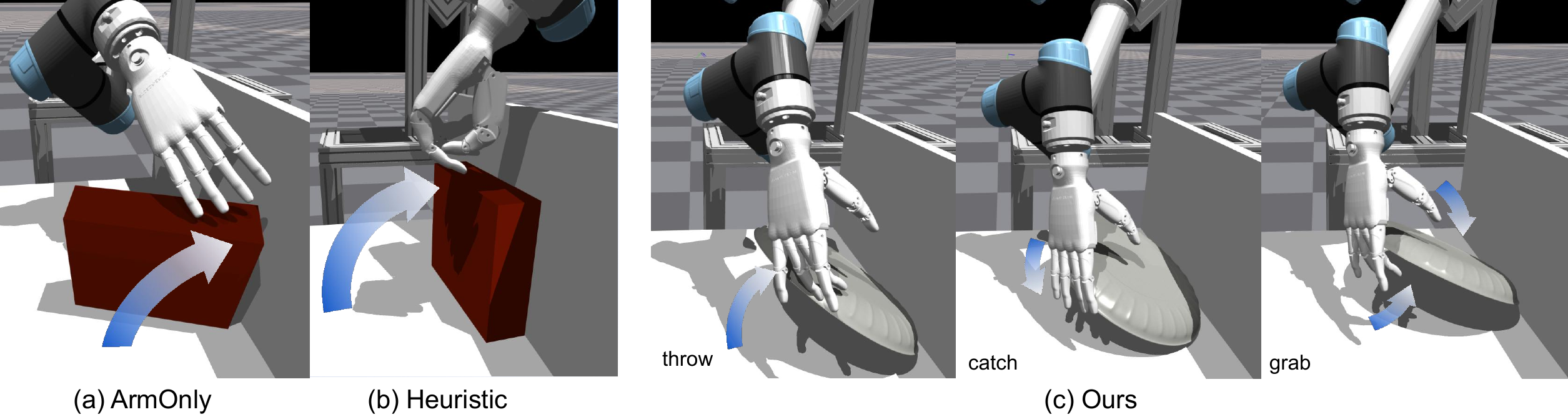}
    \caption{Comparison of \textit{Arm-Only}, \textit{Heuristic}, and \textit{Ours} for Wall task. (a) \textit{Arm-Only}. (b) \textit{Heuristic}. (c) \textit{Ours}.}
    \label{fig:re_AO_He_Ours}
\end{figure}

\textbf{Dexterous hand motion}\label{subsec:dexterous}
We compare our method with \textit{Arm-Only} and \textit{Heuristic} to validate the importance of dexterous hand motion for non-prehensile manipulation. As evidenced in Table~\ref{tab:baseline_SR}, our approach demonstrates consistent superiority over both baseline methods across all task configurations. 
The most notable performance gap emerges in the Wall task, where \textit{Arm-Only} and \textit{Heuristic} exhibit fundamental limitations due to their constrained manipulation strategies.
\textit{Arm-Only} can only pivot the object upright like Fig~\ref{fig:re_AO_He_Ours}(a), causing failure in unseen non box-shaped objects. 
\textit{Heuristic} follows a circular arc trajectory centered on the midpoint of the contact line between the corner and the object. However, it can only rotate the object to squeeze up against the wall, which prevents stable grasping (Fig~\ref{fig:re_AO_He_Ours}(b)). 
Our method overcomes these limitations through the learned dexterous hand motion. Specifically, our RL policy $\pi_{\text{wall}}$ leverages finger motions to lift and dynamically catch the object mid-air (Fig.~\ref{fig:re_AO_He_Ours}(c)), demonstrating superior dexterity. This advantage extends to FrontEdge and LeftEdge task, where our approach maintains robust performance across both seen and unseen objects.  


\textbf{Reward Design}\label{subsec:reward}
To investigate the importance of our reward design in low-level policy learning, we conduct ablation studies on two key components: (1) the motion reward (\textit{Ours w/o MR}), which guide the object toward the target pose, and (2) the stage reward mechanism (\textit{Ours w/o ST}), which dynamically adjust different reward components during training. 
As shown in Table~\ref{tab:baseline_SR}, removing the motion reward (\textit{Ours w/o MR}) leads to near zero success rates across all tasks, demonstrating that precise motion reward guidance is essential for low-level policy training.
Similarly, ablating the stage reward mechanism (\textit{Ours w/o ST}) causes a drastic performance drop, which confirms that dynamic reward adjustment is critical for smooth transitions between task stages. 
 
\textbf{Joint Finetuning}\label{subsec:joint}
To assess the effectiveness of our joint finetuning approach in enhancing the framework's generalization capability and robustness, we conducted systematic ablation studies examining both the high-level planner (\textit{Ours w/o JH}) and low-level controller (\textit{Ours w/o JL}) components. 
The experimental results Table~\ref{tab:baseline_SR} reveal that removing joint finetuning for the low-level controller results in substantial performance degradation, particularly on unseen object.
Specifically, the success rate drops approaching 40\% in these cases, clearly demonstrating that our joint finetuning approach effectively bridge the gap of the state mismatch of chaining low-level policies.
While the baseline configuration without high-level planner finetuning (\textit{Ours w/o JH}) maintains reasonable performance across all tasks, our analysis shows that incorporating target poses from successful demonstrations to refine the high-level planner yields consistent slight performance improvements. This suggests that both components of our joint finetuning strategy contribute meaningfully to the framework's overall effectiveness.


\begin{figure*}[!t]
    \centering
    \includegraphics[width=1.0\linewidth]{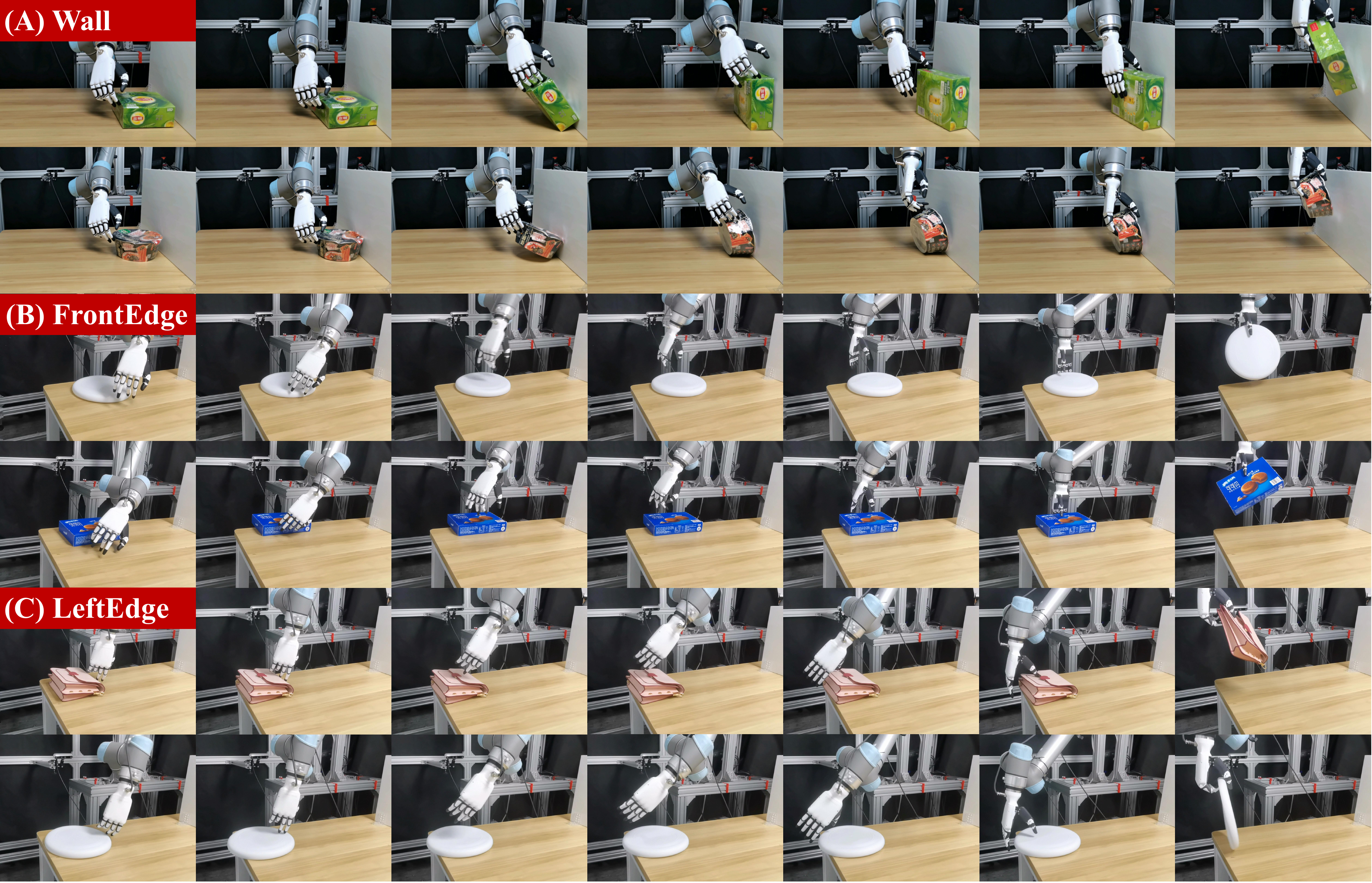}
    \caption{Real-world experiment demonstrations. The snapshots show successful executions of our framework on various objects. (a) Wall tasks. (b) FrontEdge tasks. (c) LeftEdge tasks.}
    \label{fig:exp1}
\end{figure*}

\subsection{Real-world Experiment}\label{subsec:real-world}

\begin{table}[!t]

\caption{Results for Real-world Experiments using teacher-student distillation}

\centering
\label{tab:rebuttal_realworld}
\setlength{\tabcolsep}{4pt}
\tiny
\begin{tabular}{l|ccccc}

\toprule
Size ($cm^3$)
& 16.5x15.1x6.2 
& 17.3x17.3x7.5
& 23x16.2x5
& 20.7x16.5x7
& 19x14x4  \\

\midrule
Wall  & 10/10  & 10/10  & 9/10  & 8/10  & 8/10  \\

\toprule
 Size ($cm^3$)
 & 21x13x4.4
 & 22x16x4.4 
 & 24.7x23.8x4
 & 20.7x16.5x7
 & 23.5x23.5x2.3 \\
 
\midrule
FrontEdge  & 7/10  & 9/10  & 9/10  & 6/10  & 8/10  \\

\toprule
Size ($cm^3$)
 & 21x13x4.4
 & 22x16x4.4 
 & 24.7x23.8x4
 & 20.7x16.5x7
 & 23.5x23.5x2.3 \\
\midrule
LeftEdge & 5/10  & 7/10  & 8/10  & 5/10  & 7/10  \\

\bottomrule

\end{tabular}
\vspace{-14pt}

\end{table}

\textbf{Hardware Setup} We set up the identical scenario in the real world as in the simulation, one multi-finger Inspire Hand mounted on a UR5e robot for our experiments, as shown in Figure~\ref{fig:setup}. To obtain the real-time visual observation for object pose estimation and environment state extraction, we use a single RealSense D455 depth camera. 

\textbf{Evaluation and Metric}
We evaluate our framework in real-world scenarios following the same protocols as our simulation experiments. For all tasks, objects are initially placed randomly in the center of the table. The robot then needs to execute the complete manipulation sequence: predict the target position, pushing the object to the target position near environmental contacts (wall, front or left edge) and subsequently performing the corresponding extrinsic dexterity skill to grasp it. We use the success rate (SR) as our evaluation metric, following the same criteria defined in Section~\ref{exp:SR}, where a grasp is considered successful if the object is lifted steadily above a specified height threshold.

\textbf{Object set}
To thoroughly evaluate the sim-to-real transfer capability of our framework, we conduct experiments on a diverse set of real-world objects that vary significantly in their physical properties. Our test objects, illustrated in Figure~\ref{fig:setup}, include items with different geometries, sizes, and masses. Moreover, we deliberately include several deformable objects, which present additional challenges for non-prehensile manipulation due to their changing dynamics during interaction. This diverse object set allows us to assess the robustness of our policies when dealing with physical properties that are challenging to model accurately in simulation.

\textbf{Sim-to-real performance}
As shown in Table~\ref{tab:realworld}, our policies achieves robust performance on real-world objects, with success rates exceeding 80\% for most tasks, demonstrating effective sim-to-real transfer capability. More importantly, our framework maintains high success rates even when handling objects that differ significantly from the simulation training set in terms of size and physical properties. This robust performance extends to challenging scenarios involving deformable objects, whose dynamics are particularly difficult to model accurately in simulation. These results highlight the strong generalization capability and robustness of our trained policies, successfully bridging the reality gap in complex manipulation tasks. The detailed visualization of our real-world experiments are shown in Figure~\ref{fig:exp1}.

\section{Limitation} 
\label{sec:limitation}
There are several limitations of our work.
(1) \textit{Limited operation space}. Our current implementation relies on a fixed-base robot arm, which constrains the operational workspace to a corner region of the table. This limitation restricts the position generalizability of our policies, as the robot cannot access objects placed beyond its reachable workspace. A potential solution would be to integrate our framework with a mobile manipulator, which would enable extrinsic dexterity capabilities across larger working areas.
(2) \textit{Clutter scene generalizability}. All of our experiments are conducted on a clean table which only contains wall or target objects. However, a common table in daily life is usually filled with cluttered objects, which not only introduces additional obstacles for manipulation but also provides potential new external contacts for extrinsic dexterity. Future work could focus on enhancing both our prediction model and pushing policy to enable robust object repositioning in cluttered environments, effectively identifying and utilizing suitable positions for extrinsic dexterity among obstacles.  



\section{Conclusion} 
\label{sec:conclusion}

In this work, we investigate the challenging problem of manipulating ungraspable objects using extrinsic dexterity with a multi-finger hand. Inspired by human's ability to leverage environmental features like walls and edges, we present a hierarchical framework that combines strategic planning with dexterous manipulation skills. Our framework features a high-level planner that intelligently selects optimal external contacts and predicts target positions, coupled with a low-level controller that executes precise non-prehensile manipulation skills.
Through extensive experiments in simulation, we demonstrate our framework's superior performance across different external contacts and various objects. The results show that our approach successfully addresses the key challenges in extrinsic dexterity by three factors: strategic object repositioning, dynamic contact interactions, and precise manipulation control. Moreover, the successful transfer of our policies from simulation to a real-world robot system validates the practical applicability of our method, bridging the gap between simulation and reality in contact-rich manipulation tasks.

\bibliographystyle{unsrt}
\bibliography{references}

\begin{thebibliography}{10}

\bibitem{chen2024springgrasp}
Sirui Chen, Jeannette Bohg, and C~Karen Liu.
\newblock Springgrasp: An optimization pipeline for robust and compliant dexterous pre-grasp synthesis.
\newblock {\em arXiv preprint arXiv:2404.13532}, 2024.

\bibitem{mordatch2012contact}
Igor Mordatch, Zoran Popovi{\'c}, and Emanuel Todorov.
\newblock Contact-invariant optimization for hand manipulation.
\newblock In {\em Proceedings of the ACM SIGGRAPH/Eurographics symposium on computer animation}, pages 137--144, 2012.

\bibitem{chen2022learning}
Zoey~Qiuyu Chen, Karl Van~Wyk, Yu-Wei Chao, Wei Yang, Arsalan Mousavian, Abhishek Gupta, and Dieter Fox.
\newblock Learning robust real-world dexterous grasping policies via implicit shape augmentation.
\newblock {\em arXiv preprint arXiv:2210.13638}, 2022.

\bibitem{shaw2023videodex}
Kenneth Shaw, Shikhar Bahl, and Deepak Pathak.
\newblock Videodex: Learning dexterity from internet videos.
\newblock In {\em Conference on Robot Learning}, pages 654--665. PMLR, 2023.

\bibitem{wang2024dexcap}
Chen Wang, Haochen Shi, Weizhuo Wang, Ruohan Zhang, Li~Fei-Fei, and C~Karen Liu.
\newblock Dexcap: Scalable and portable mocap data collection system for dexterous manipulation.
\newblock {\em arXiv preprint arXiv:2403.07788}, 2024.

\bibitem{openai2019solving}
OpenAI, Ilge Akkaya, Marcin Andrychowicz, Maciek Chociej, Mateusz Litwin, Bob McGrew, Arthur Petron, Alex Paino, Matthias Plappert, Glenn Powell, Raphael Ribas, Jonas Schneider, Nikolas Tezak, Jerry Tworek, Peter Welinder, Lilian Weng, Qiming Yuan, Wojciech Zaremba, and Lei Zhang.
\newblock Solving rubik's cube with a robot hand.
\newblock {\em CoRR}, abs/1910.07113, 2019.

\bibitem{yang2024anyrotate}
Max Yang, Chenghua Lu, Alex Church, Yijiong Lin, Chris Ford, Haoran Li, Efi Psomopoulou, David~AW Barton, and Nathan~F Lepora.
\newblock Anyrotate: Gravity-invariant in-hand object rotation with sim-to-real touch.
\newblock {\em arXiv preprint arXiv:2405.07391}, 2024.

\bibitem{pitz2023dextrous}
Johannes Pitz, Lennart R{\"o}stel, Leon Sievers, and Berthold B{\"a}uml.
\newblock Dextrous tactile in-hand manipulation using a modular reinforcement learning architecture.
\newblock In {\em 2023 IEEE International Conference on Robotics and Automation (ICRA)}, pages 1852--1858. IEEE, 2023.

\bibitem{handa2022dextreme}
Ankur Handa, Arthur Allshire, Viktor Makoviychuk, Aleksei Petrenko, Ritvik Singh, Jingzhou Liu, Denys Makoviichuk, Karl Van~Wyk, Alexander Zhurkevich, Balakumar Sundaralingam, et~al.
\newblock Dextreme: Transfer of agile in-hand manipulation from simulation to reality.
\newblock {\em arXiv preprint arXiv:2210.13702}, 2022.

\bibitem{chen2021system}
Tao Chen, Jie Xu, and Pulkit Agrawal.
\newblock A system for general in-hand object re-orientation.
\newblock {\em Conference on Robot Learning}, 2021.

\bibitem{zhou2023learning}
Wenxuan Zhou and David Held.
\newblock Learning to grasp the ungraspable with emergent extrinsic dexterity.
\newblock In {\em Conference on Robot Learning}, pages 150--160. PMLR, 2023.

\bibitem{chen2023synthesizing}
Sirui Chen, Albert Wu, and C~Karen Liu.
\newblock Synthesizing dexterous nonprehensile pregrasp for ungraspable objects.
\newblock In {\em ACM SIGGRAPH 2023 Conference Proceedings}, pages 1--10, 2023.

\bibitem{bai2014dexterous}
Yunfei Bai and C~Karen Liu.
\newblock Dexterous manipulation using both palm and fingers.
\newblock In {\em 2014 IEEE International Conference on Robotics and Automation (ICRA)}, pages 1560--1565. IEEE, 2014.

\bibitem{kumar2014real}
Vikash Kumar, Yuval Tassa, Tom Erez, and Emanuel Todorov.
\newblock Real-time behaviour synthesis for dynamic hand-manipulation.
\newblock In {\em 2014 IEEE International Conference on Robotics and Automation (ICRA)}, pages 6808--6815. IEEE, 2014.

\bibitem{mandikal2021learning}
Priyanka Mandikal and Kristen Grauman.
\newblock Learning dexterous grasping with object-centric visual affordances.
\newblock In {\em 2021 IEEE international conference on robotics and automation (ICRA)}, pages 6169--6176. IEEE, 2021.

\bibitem{chen2022dextransfer}
Zoey~Qiuyu Chen, Karl Van~Wyk, Yu-Wei Chao, Wei Yang, Arsalan Mousavian, Abhishek Gupta, and Dieter Fox.
\newblock Dextransfer: Real world multi-fingered dexterous grasping with minimal human demonstrations.
\newblock {\em arXiv preprint arXiv:2209.14284}, 2022.

\bibitem{rajeswaran2017learning}
Aravind Rajeswaran, Vikash Kumar, Abhishek Gupta, Giulia Vezzani, John Schulman, Emanuel Todorov, and Sergey Levine.
\newblock Learning complex dexterous manipulation with deep reinforcement learning and demonstrations.
\newblock {\em arXiv preprint arXiv:1709.10087}, 2017.

\bibitem{radosavovic2021state}
Ilija Radosavovic, Xiaolong Wang, Lerrel Pinto, and Jitendra Malik.
\newblock State-only imitation learning for dexterous manipulation.
\newblock In {\em 2021 IEEE/RSJ International Conference on Intelligent Robots and Systems (IROS)}, pages 7865--7871. IEEE, 2021.

\bibitem{arunachalam2023holo}
Sridhar~Pandian Arunachalam, Irmak G{\"u}zey, Soumith Chintala, and Lerrel Pinto.
\newblock Holo-dex: Teaching dexterity with immersive mixed reality.
\newblock In {\em 2023 IEEE International Conference on Robotics and Automation (ICRA)}, pages 5962--5969. IEEE, 2023.

\bibitem{guzey2023dexterity}
Irmak Guzey, Ben Evans, Soumith Chintala, and Lerrel Pinto.
\newblock Dexterity from touch: Self-supervised pre-training of tactile representations with robotic play.
\newblock {\em arXiv preprint arXiv:2303.12076}, 2023.

\bibitem{handa2020dexpilot}
Ankur Handa, Karl Van~Wyk, Wei Yang, Jacky Liang, Yu-Wei Chao, Qian Wan, Stan Birchfield, Nathan Ratliff, and Dieter Fox.
\newblock Dexpilot: Vision-based teleoperation of dexterous robotic hand-arm system.
\newblock In {\em 2020 IEEE International Conference on Robotics and Automation (ICRA)}, pages 9164--9170. IEEE, 2020.

\bibitem{sivakumar2022robotic}
Aravind Sivakumar, Kenneth Shaw, and Deepak Pathak.
\newblock Robotic telekinesis: Learning a robotic hand imitator by watching humans on youtube.
\newblock {\em arXiv preprint arXiv:2202.10448}, 2022.

\bibitem{qin2023anyteleop}
Yuzhe Qin, Wei Yang, Binghao Huang, Karl Van~Wyk, Hao Su, Xiaolong Wang, Yu-Wei Chao, and Dietor Fox.
\newblock Anyteleop: A general vision-based dexterous robot arm-hand teleoperation system.
\newblock {\em arXiv preprint arXiv:2307.04577}, 2023.

\bibitem{qin2022dexmv}
Yuzhe Qin, Yueh-Hua Wu, Shaowei Liu, Hanwen Jiang, Ruihan Yang, Yang Fu, and Xiaolong Wang.
\newblock Dexmv: Imitation learning for dexterous manipulation from human videos.
\newblock In {\em European Conference on Computer Vision}, pages 570--587. Springer, 2022.

\bibitem{cui2022play}
Zichen~Jeff Cui, Yibin Wang, Nur Muhammad~Mahi Shafiullah, and Lerrel Pinto.
\newblock From play to policy: Conditional behavior generation from uncurated robot data.
\newblock {\em arXiv e-prints}, pages arXiv--2210, 2022.

\bibitem{haldar2023teach}
Siddhant Haldar, Jyothish Pari, Anant Rai, and Lerrel Pinto.
\newblock Teach a robot to fish: Versatile imitation from one minute of demonstrations.
\newblock {\em arXiv preprint arXiv:2303.01497}, 2023.

\bibitem{qin2022one}
Yuzhe Qin, Hao Su, and Xiaolong Wang.
\newblock From one hand to multiple hands: Imitation learning for dexterous manipulation from single-camera teleoperation.
\newblock {\em IEEE Robotics and Automation Letters}, 7(4):10873--10881, 2022.

\bibitem{arunachalam2022dexterous}
Sridhar~Pandian Arunachalam, Sneha Silwal, Ben Evans, and Lerrel Pinto.
\newblock Dexterous imitation made easy: A learning-based framework for efficient dexterous manipulation.
\newblock {\em arXiv preprint arXiv:2203.13251}, 2022.

\bibitem{guzey2023see}
Irmak Guzey, Yinlong Dai, Ben Evans, Soumith Chintala, and Lerrel Pinto.
\newblock See to touch: Learning tactile dexterity through visual incentives.
\newblock {\em arXiv preprint arXiv:2309.12300}, 2023.

\bibitem{lin2024learning}
Toru Lin, Yu~Zhang, Qiyang Li, Haozhi Qi, Brent Yi, Sergey Levine, and Jitendra Malik.
\newblock Learning visuotactile skills with two multifingered hands.
\newblock {\em arXiv preprint arXiv:2404.16823}, 2024.

\bibitem{huang20243dvitac}
Binghao Huang, Yixuan Wang, Xinyi Yang, Yiyue Luo, and Yunzhu Li.
\newblock 3d vitac:learning fine-grained manipulation with visuo-tactile sensing.
\newblock In {\em Proceedings of Robotics: Conference on Robot Learning(CoRL)}, 2024.

\bibitem{chen2022visual}
Tao Chen, Megha Tippur, Siyang Wu, Vikash Kumar, Edward Adelson, and Pulkit Agrawal.
\newblock Visual dexterity: In-hand dexterous manipulation from depth.
\newblock {\em arXiv preprint arXiv:2211.11744}, 2022.

\bibitem{yin2023rotating}
Zhao-Heng Yin, Binghao Huang, Yuzhe Qin, Qifeng Chen, and Xiaolong Wang.
\newblock Rotating without seeing: Towards in-hand dexterity through touch.
\newblock {\em arXiv preprint arXiv:2303.10880}, 2023.

\bibitem{qi2023general}
Haozhi Qi, Brent Yi, Sudharshan Suresh, Mike Lambeta, Yi~Ma, Roberto Calandra, and Jitendra Malik.
\newblock General in-hand object rotation with vision and touch.
\newblock In {\em Conference on Robot Learning}, pages 2549--2564. PMLR, 2023.

\bibitem{qi2023hand}
Haozhi Qi, Ashish Kumar, Roberto Calandra, Yi~Ma, and Jitendra Malik.
\newblock In-hand object rotation via rapid motor adaptation.
\newblock In {\em Conference on Robot Learning}, pages 1722--1732. PMLR, 2023.

\bibitem{dasari2023learning}
Sudeep Dasari, Abhinav Gupta, and Vikash Kumar.
\newblock Learning dexterous manipulation from exemplar object trajectories and pre-grasps.
\newblock In {\em 2023 IEEE International Conference on Robotics and Automation (ICRA)}, pages 3889--3896. IEEE, 2023.

\bibitem{khandate2023sampling}
Gagan Khandate, Siqi Shang, Eric~T Chang, Tristan~Luca Saidi, Yang Liu, Seth~Matthew Dennis, Johnson Adams, and Matei Ciocarlie.
\newblock Sampling-based exploration for reinforcement learning of dexterous manipulation.
\newblock {\em arXiv preprint arXiv:2303.03486}, 2023.

\bibitem{huang2023dynamic}
Binghao Huang, Yuanpei Chen, Tianyu Wang, Yuzhe Qin, Yaodong Yang, Nikolay Atanasov, and Xiaolong Wang.
\newblock Dynamic handover: Throw and catch with bimanual hands.
\newblock {\em arXiv preprint arXiv:2309.05655}, 2023.

\bibitem{lin2024twisting}
Toru Lin, Zhao-Heng Yin, Haozhi Qi, Pieter Abbeel, and Jitendra Malik.
\newblock Twisting lids off with two hands.
\newblock {\em arXiv preprint arXiv:2403.02338}, 2024.

\bibitem{chen2022towards}
Yuanpei Chen, Tianhao Wu, Shengjie Wang, Xidong Feng, Jiechuan Jiang, Zongqing Lu, Stephen McAleer, Hao Dong, Song-Chun Zhu, and Yaodong Yang.
\newblock Towards human-level bimanual dexterous manipulation with reinforcement learning.
\newblock {\em Advances in Neural Information Processing Systems}, 35:5150--5163, 2022.

\bibitem{ding2024preafford}
Kairui Ding, Boyuan Chen, Ruihai Wu, Yuyang Li, Zongzheng Zhang, Huan-ang Gao, Siqi Li, Guyue Zhou, Yixin Zhu, Hao Dong, et~al.
\newblock Preafford: Universal affordance-based pre-grasping for diverse objects and environments.
\newblock In {\em 2024 IEEE/RSJ International Conference on Intelligent Robots and Systems (IROS)}, pages 7278--7285. IEEE, 2024.

\bibitem{chen2023sequential}
Yuanpei Chen, Chen Wang, Li~Fei-Fei, and C~Karen Liu.
\newblock Sequential dexterity: Chaining dexterous policies for long-horizon manipulation.
\newblock {\em arXiv preprint arXiv:2309.00987}, 2023.

\bibitem{zakka2023robopianist}
Kevin Zakka, Laura Smith, Nimrod Gileadi, Taylor Howell, Xue~Bin Peng, Sumeet Singh, Yuval Tassa, Pete Florence, Andy Zeng, and Pieter Abbeel.
\newblock Robopianist: A benchmark for high-dimensional robot control.
\newblock {\em arXiv preprint arXiv:2304.04150}, 2023.

\bibitem{dafle2014extrinsic}
Nikhil~Chavan Dafle, Alberto Rodriguez, Robert Paolini, Bowei Tang, Siddhartha~S Srinivasa, Michael Erdmann, Matthew~T Mason, Ivan Lundberg, Harald Staab, and Thomas Fuhlbrigge.
\newblock Extrinsic dexterity: In-hand manipulation with external forces.
\newblock In {\em 2014 IEEE International Conference on Robotics and Automation (ICRA)}, pages 1578--1585. IEEE, 2014.

\bibitem{ma2024dexdiff}
Chengzhong Ma, Houxue Yang, Hanbo Zhang, Zeyang Liu, Chao Zhao, Jian Tang, Xuguang Lan, and Nanning Zheng.
\newblock Dexdiff: Towards extrinsic dexterity manipulation of ungraspable objects in unrestricted environments.
\newblock {\em arXiv preprint arXiv:2409.05493}, 2024.

\bibitem{stepputtis2018extrinsic}
Simon Stepputtis, Yezhou Yang, and Heni~Ben Amor.
\newblock Extrinsic dexterity through active slip control using deep predictive models.
\newblock In {\em 2018 IEEE International Conference on Robotics and Automation (ICRA)}, pages 3180--3185. IEEE, 2018.

\bibitem{dong2023robotic}
Yi~Dong, Jinjun Duan, Yangjun Liu, Zhendong Dai, and Poramate Manoonpong.
\newblock Robotic shoe packaging strategies based on a single soft-gripper system and extrinsic resources.
\newblock In {\em 2023 International Conference on Advanced Robotics and Mechatronics (ICARM)}, pages 469--475. IEEE, 2023.

\bibitem{ha2022flingbot}
Huy Ha and Shuran Song.
\newblock Flingbot: The unreasonable effectiveness of dynamic manipulation for cloth unfolding.
\newblock In {\em Conference on Robot Learning}, pages 24--33. PMLR, 2022.

\bibitem{wu2024unidexfpm}
Tianhao Wu, Yunchong Gan, Mingdong Wu, Jingbo Cheng, Yaodong Yang, Yixin Zhu, and Hao Dong.
\newblock Unidexfpm: Universal dexterous functional pre-grasp manipulation via diffusion policy.
\newblock {\em arXiv preprint arXiv:2403.12421}, 2024.

\bibitem{schulman2017proximal}
John Schulman, Filip Wolski, Prafulla Dhariwal, Alec Radford, and Oleg Klimov.
\newblock Proximal policy optimization algorithms.
\newblock {\em arXiv preprint arXiv:1707.06347}, 2017.

\bibitem{konidaris2009skill}
George Konidaris and Andrew Barto.
\newblock Skill chaining: Skill discovery in continuous domains.
\newblock In {\em the Multidisciplinary Symposium on Reinforcement Learning, Montreal, Canada}, 2009.

\bibitem{sutton1999between}
Richard~S Sutton, Doina Precup, and Satinder Singh.
\newblock Between mdps and semi-mdps: A framework for temporal abstraction in reinforcement learning.
\newblock {\em Artificial intelligence}, 112(1-2):181--211, 1999.

\bibitem{kirillov2023segment}
Alexander Kirillov, Eric Mintun, Nikhila Ravi, Hanzi Mao, Chloe Rolland, Laura Gustafson, Tete Xiao, Spencer Whitehead, Alexander~C Berg, Wan-Yen Lo, et~al.
\newblock Segment anything.
\newblock In {\em Proceedings of the IEEE/CVF International Conference on Computer Vision}, pages 4015--4026, 2023.

\bibitem{wen2023foundationpose}
Bowen Wen, Wei Yang, Jan Kautz, and Stan Birchfield.
\newblock Foundationpose: Unified 6d pose estimation and tracking of novel objects.
\newblock {\em arXiv preprint arXiv:2312.08344}, 2023.

\end{thebibliography}

\newpage
\appendix


\section*{Appendix}









\section{Hyperparameters of the PPO}
\begin{table}[thbp]
\caption{Hyperparameters of PPO.}
    \centering
    \begin{tabular}{c|c}
    \toprule
    Hyperparameters & Value  \\\hline
    Num mini-batches & 4 \\
    Num opt-epochs & 5 \\
    Num episode-length & 8\\\hline
    Hidden size & [1024, 512, 256] \\
    Clip range & 0.2\\
    Max grad norm & 1\\
    Learning rate & 5e-4 \\
    Discount ($\gamma$) & 0.99\\
    GAE lambda ($\lambda$) & 0.95\\
    Init noise std & 1.0\\\hline
    Desired kl & 0.008 \\
    Ent-coef & 0\\
    \bottomrule
    \end{tabular}
    
    \label{tab:hyper_ppo}
\end{table}
Table~\ref{tab:hyper_ppo} shows the hyperparameters of the PPO.

\subsection{Reward Design}\label{app:reward_design}
\begin{figure}[!ht]
    \centering
    \includegraphics[width=0.95\linewidth]{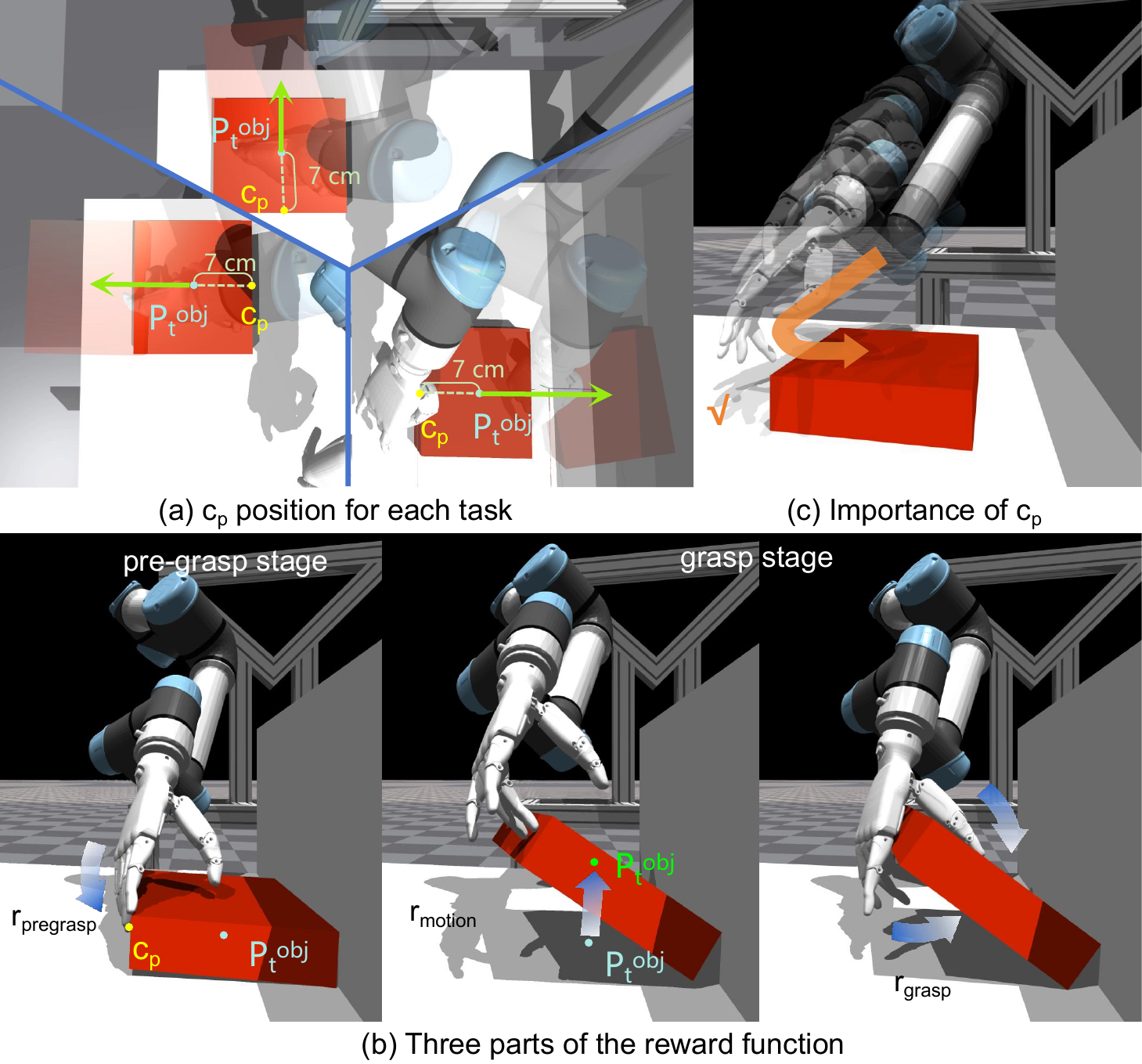}
    \caption{\textbf{Illustration for reward design.} (a) $c_p$ position for each task. (b) Three parts of the reward function. (c) Importance of $c_p$. For $\pi_{\textrm{wall}}$, }  
    \label{fig:appendix_pic}
\end{figure}

Instead of being manually designed for each object, the contact point $c_p \in \mathbb{R}^{3}$ is defined as a fixed spatial offset from the center point of all objects, maintaining a constant 7 cm distance along the object's width axis as show in Figure~\ref{fig:appendix_pic}(a) without manually designed for each object. Given an object with its center point $P_t^{obj} = (x, y, z)$, the contact point is computed as
\begin{equation}
c_p = P_t^{obj} + d\cdot\hat{w}
\end{equation}
where $d = 7cm$ represents the fixed offset distance and $\hat{w}$ denotes the unit vector along the object's width dimension.


In  Equation~\ref{equation:rew} , we divide the reward function into three parts: $r_{\textrm{motion}}$, $r_{\textrm{pregrasp}}$, and $r_{\textrm{grasp}}$.
$r_{\textrm{motion}}$ guides the policy toward its ultimate goal and remains active throughout the entire task execution.
$r_{\textrm{pregrasp}}$ encourages the dexterous hand to move towards the object for pre-grasp following the trajectory we expect (first moving towards $c_p$, and then moving towards the object center $P_t^{obj}$). 
$r_{\textrm{grasp}}$ is designed to facilitate successful object grasping after pre-grasp.


For $\pi_{\textrm{wall}}$, we expect that the hand first approaches the object from its side guided by $r_{\textrm{pregrasp}}$, and then grasp it between the thumb and other fingers guided by $r_{\textrm{grasp}}$. Thus, we set $P(a)=1, P(b)=0$ in the pre-grasp stage and switch to $P(a)=0, P(b)=1$ in the grasp stage. The switch occurs when $r_{\textrm{pregrasp}} < 3cm$ which indicates sufficient proximity between the middle finger $F_t^{f,3}$ and $c_p$. $r_{\textrm{motion}}$ continuously compute the distance between the object $P_t^{obj}$ and a point $P_t^{target} = P_t^{obj} + [0, 0, 10](cm)$. Notice that this point locates above the object which is designed to guide the robot to rotate up the object.


For $\pi_{\textrm{edge}}$, the object is intentionally positioned to expose a graspable side at the table edge. Therefore, the pre-grasp phase is eliminated ($P(a)\equiv0$) since moving the finger to $c_p$ becomes unnecessary. Besides, the object translation in the horizontal plane is no longer required, allowing the policy to focus exclusively on vertical finger coordination. The motion reward is formulated as:
\begin{equation}
r_{\textrm{motion}} = -\sum_{i=1}^5 |F_t^{f,i} - P_t^{target}|
\end{equation}
where target heights are:
\begin{equation}
P_t^{target} = 
\begin{cases}
P_t^{obj} + [0, 0, 0.15]\text{cm}, & \text{for thumb finger} \\
P_t^{obj} - [0, 0, 0.05]\text{cm}, & \text{for other fingers}
\end{cases}
\end{equation}
We set $P(b)=1$ if the middle finger moves below the object to activate the grasp stage. Otherwise, we maintain $P(b)=0$.


For $\pi_{\textrm{push}}$, stage training and grasping are unnecessary, resulting in $P(a)\equiv1$ and $P(b)\equiv0$. $r_{\textrm{pregrasp}}$ guides the hand to push the object from its side, and $r_{\textrm{motion}}$ narrows the gap between the object $P_t^{obj}$ and $P_t^{target}$ predicted by our high-level planner.


\subsection{Sim-to-real Details}
\label{app:sim2real}
\textbf{Teacher-student distillation.} 
We collect 1000 demonstration trajectories with the teacher RL policies in simulation for each task. Here we manually design some rules to remove the unnatural or dangerous behaviors which emerge from exploiting the simulator dynamics but don't transfer well to real-world. We use a transformer-based network to imitate the curated demonstration. The network architecture is as followed:

The network takes as input a sequence of concatenated state observations (dimension: 13) spanning 10 historical frames. An initial feature extraction module processes each frame independently through two linear layers (128 and 512 units respectively), each followed by ReLU activation and layer normalization. The extracted features are augmented with learnable positional encodings to preserve temporal information. The temporal dynamics are modeled through a 3-layer transformer encoder (dmodel=512, nheads=2), where each layer contains: Multi-head self-attention for capturing frame dependencies; Position-wise feed-forward network; Residual connections and layer normalization. Following the transformer encoder, we employ global average pooling across the temporal dimension and process the features through two residual blocks for enhanced representation learning. The final action predictor consists of a carefully designed MLP with progressively decreasing layer widths (256 → 128 units), each followed by ReLU activation, layer normalization, and dropout (p=0.1). The network outputs 8 consecutive action frames (dimension: 12 per frame) through a tanh-activated linear layer, ensuring actions remain within valid bounds.

We supervise the output action $\mathbf{a}_{\text{pred}}$ with negative log product loss with L2 regularization:

\begin{small}
\begin{equation}
\mathcal{L}(\mathbf{a}_{\text{pred}}, \mathbf{a}_{\text{gt}}) = -\sum_{i=1}^{N} \log\left(1 - |\mathbf{a}_{\text{pred}}^{(i)} - \mathbf{a}_{\text{gt}}^{(i)}|1\right) + \lambda||\mathbf{a}_{\text{pred}}||_2^2
\end{equation}
\end{small}

\textbf{Digital twin.} 
\begin{table}[!ht]

\caption{Results for the real-world experiments}

\centering
\scriptsize
\label{tab:realworld}
\setlength{\tabcolsep}{1.5mm}
\begin{tabular}{l|cccccc}

\toprule
& box-w1 & box-w2  & box-w3 & bag-w1 & container & handbag\\

\midrule
Wall  & 6/10 & 8/10 &  7/10 & 9/10 & 9/10 & 9/10 \\

\toprule
 & box-e1 & box-e2  & bag-e1 & bag-e2 & plate & handbag\\

\midrule
Edge    & 10/10   & 7/10       &  10/10 & 8/10 & 5/10 & 9/10  \\

\bottomrule

\end{tabular}

\end{table}
Before leveraging the teacher-student 
distillation, we achieve zero-shot sim-to-real transfer by implementing a digital twin framework that bridges our simulation policy with the real dexterous arm-hand system. The framework operates through two parallel threads that enable real-time asynchronous communication between simulation and real-world environments. The real-world thread continuously collects observations, including arm-hand proprioception and object pose information. Meanwhile, the simulation thread processes these observations to generate control actions, executes them in simulation and uses the resulting joint angles as target joint angles for PD control in the real robot system. The simulation environment is continuously synchronized with real-world by updating robot joint angles and object poses from real-world observations. Through the constant synchronization of simulation and real-world, we evaluate our RL polices following the similar setup as mentioned in Subsection ~\ref{subsec:real-world}.
The results in Table~\ref{tab:realworld} shows that our digital twin framework achieves robust performance on real-world objects, with success rates exceeding 80\% for most objects.

\end{document}